\documentclass[preprint,12pt,number,3p]{elsarticle}
\usepackage{amssymb}
\usepackage{multirow}
\usepackage{amsmath}
\usepackage{setspace}
\usepackage[table]{xcolor}
\usepackage{url}
\usepackage{colortbl}
\usepackage{dirtree}
\usepackage{dblfloatfix}
\usepackage{array}

\journal{}

\begin{document}

\begin{frontmatter}

%Current Suggested Reviewers List
%Eirik Albrechtsen eirik.albrechtsen@ntnu.no Norges teknisk-naturvitenskapelige universitet
%Alistair Gibb a.g.gibb@lboro.ac.uk Loughborough University
%Helen Lingard helen.lingard@rmit.edu.au

\title{Causal factors discovering from Chinese construction accident cases}
\author[rvt]{Zi-jian Ni\corref{cor1}}
\ead{nizijian@hotmail.com}
%\author[rvt]{Ning Wang }%wn@dlut.edu.cn
%
%\author[focal]{Zi-jun Qie} %qiezijun@dlut.edu.cn
%
%\author[els]{Shuo Cao}%caoshuo@dlut.edu.cn
%
\author[rvt]{Wei Liu}%2535254162@qq.com
\cortext[cor1]{Corresponding author}

\address[rvt]{
Faculty of Economics and Management, Dalian University of Technology\\
2 Ling Gong Rd., Dalian 116024, Liaoning, P. R. China
}
%\address[focal]{
%School of Foreign Languages, Dalian University of Technology\\
%}
%\address[els]{
%Department of Public Administration, Dalian University of Technology\\
%2 Ling Gong Rd., Dalian 116024, Liaoning, P. R. China
%}

\begin{abstract}
In China, construction accidents have killed more people than any other industry since 2012.
The factors which led to the accident have complex interaction.
Real data about accidents is the key to reveal the mechanism among these factors.
But the data from the questionnaire and interview has inherent defects.
Many behaviors that impact safety are illegal.
In China, most of the cases are from accident investigation reports.
Finding out the cause of the accident and liability affirmation are the core of incident investigation reports.
So the truth of some answers from the respondents is doubtful.
With a series of NLP technologies, in this paper, causal factors of construction accidents are extracted and organized from Chinese incident case texts.
Finally, three kinds of neglected causal factors are discovered after data analysis.
\end{abstract}

%The factors which led to the accident have complex interaction. Real data about accidents is the key to reveal the mechanism among these factors. But the data from the questionnaire and interview has inherent defects. Many behaviors that impact safety are illegal. In China, most of the cases are from accident investigation reports. Finding out the cause of the accident and liability affirmation are the core of incident investigation reports. So the truth of some answers from the respondents is doubtful. With a series of NLP technologies, in this paper, causal factors of construction accidents are extracted and organized from Chinese incident case texts. Finally, three kinds of neglected causal factors are discovered after data analysis.


\end{frontmatter}

\section{Introduction}
China, as the largest construction market in the world, its value of construction output was about 24.8 trillion Yuan in 2019.
Concerning safety in the construction industry, it is still challenging today \citep{tam2004identifying}.
The death toll reached 1152 in 2003 and then fell for 11 consecutive years.
With the holistic improvement of the occupational health and safety management system of the country, however, accidents in the construction industry have killed more people than in coal mines since 2012.
In 2019, construction deaths on the job were 904, which ranked the first in all types of industrial accidents.
Many studies hold that construction is one of the most dangerous industries due to the complicated and multicausal factor of accidents on project sites \citep{suraji2001development, mohseni2015assessment}.

In \citep{khanzode2012occupational}, accident causation theories were divided into four generations: accident proneness theories, domino theories, injury epidemiology models, and system theories.
In the last generation, occupational safety is impacted by factors in different levels that have complex interactions.
Further, two kinds of elements are analyzed in the construction accident system model.
One is the factors influencing safety performance,  which is called the risk factor.
And the other is the causal factor.
As the name implies, they resulted in the accident.

Generally, the system model about risk factors is based on the empirical method.
The whole research begins with statements or hypotheses.
After data collection from the questionnaire and interview, whether a hypothesis is supported or not depends on the appropriate statistical formula.
All kinds of specific aspects of construction safety have been discussed in this methodology.
Thirteen main risk factors from 55 papers are summarized in a useful review \citep{mohammadi2018factors}.
In construction accident analysis, there is an essential weakness of this kind of empirical research.
Many behaviors that impact safety is illegal.
So the truth of some answers from the respondents in the questionnaire and interview are doubtful.

Moreover, the unsafety does not equal to the accident.
Revealing the causality of accidents is essential to distinguish between factors that require some action or not \citep{gibb2014construction}.
The research shows that causes of accidents vary substantially between industries \citep{williamson1996industry}.
Most causal models of construction accident \citep{suraji2001development, hide2003causal, mitropoulos2005systems, hale2012developing} originated from systematic and holistic thinking about accidents.
But not all of them have been validated by sufficient real accident data.
In work \citep{hale2012developing}, for example, only a small sample of fatal accidents (26 in total 211 accidents cases) was used to understand underlying causes.
Another example is causal factors were divided into the proximal and distal in the \citep{suraji2001development}.
But because of the limitations of the accident data available, only the proximal factors are validated \citep{suraji2001development}.
The ConAC (Construction Accident Causality) framework \citep{hide2003causal} was verified \citep{gibb2014construction, winge2019causal} and applied \citep{behm2013application} a couple of times.
But at the same time, analyzing data is the cost.
For extraction data from 84000 words, this study engaged four analysts
 \citep{winge2019causal}.
The consistency of criteria for extracting information is still problematic, even if you can hire more skilled professionals.
As a result, for analyzing construction accidents, real data is the key.

Not only in the field of construction, but it is also hard to collect data of accidents in other industries.
The reason is that it is impossible to conduct reproducible incident experiments like other disciplines.
Past accident analysis and learning (PAL) is always one of two pillars on which the edifice of occupational safety research \citep{abdolhamidzadeh2011domino}.
For PAL, accident cases are one of the most important sources \citep{tauseef2011development}.
In China, most of the cases are from accident investigation reports \citep{Housing2019}.
Finding out the cause of the accident and liability affirmation are the core of incident investigation reports. \citep{PeoplesRepublicofChina2007}.
Including illegal acts, in other words, causal factors of every accident can be found in these documents.
NLP (Natural Language Processing) can assist people in improving the performance of analyzing the unstructured text.
In this paper, causal factors of construction accidents will be extracted from the free text in Chinese with Automatic Keyphrase Extraction (AKE) \citep{merrouni2016automatic}.
AKE includes a series of NLP technologies and will be discussed in section \ref{sec:extract}.
Furthermore, not only for incidents of construction, we believe that our framework for the extraction can be used in other industry accident case text in Chinese.

For evaluating the necessity and sufficiency of causal factors in data sets, all valid accident data in a \texttt{short-term} was input into various algorithms to get the correlation.
Because our Chinese cases are typical incidents for an extended period (more than 25 years), the holistic causal model can not be proposed in this paper.
But due to more accurate information being extracted and summarized, some neglected causal factors will be revealed.
In the meanwhile, empirical studies may be inspired by these real accident data also.
Finally, the organized data will be shared online for further studies  \footnote{\url{https://github.com/liuwei-965/Digital-management-of-Chinese-accident-cases}}.

The rest of this paper is structured as follows:
The data source and the case text structure will be introduced in section \ref{sec:case_text}.
A framework for extraction causal factors from texts will be proposed in section \ref{sec:extract}.
In section \ref{sec:discovery}, the role mismatch and the other two neglected factors will be discussed.

\section{Accident causes in Chinese accident cases}\label{sec:case_text}
\begin{figure}[ht]
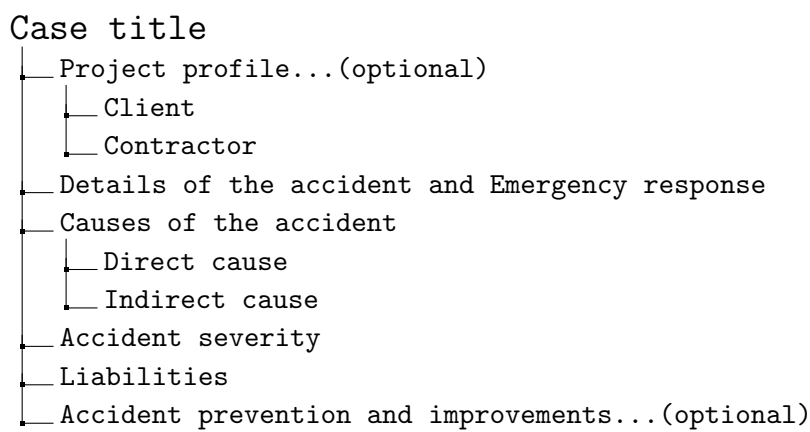

 \small
\dirtree{%
.1 \large{Case title}.
.2 \textbf{Project profile}\ldots{}(optional).
.3 Client.
.3 Contractor.
.2 \textbf{Details of the accident and Emergency response}.
.2 \textbf{Causes of the accident}.
.3 Direct cause.
.3 Indirect cause.
.2 \textbf{Accident severity}.
.2 \textbf{Liabilities}.
.2 \textbf{Accident prevention and improvements}\ldots{}(optional).
}
\caption{The structure of Chinese construction accident cases.}
\label{tree:structure}
\end{figure}

In our study, 267 typical construction accident cases are all from esafety.cn, which is the information platform of the Ministry of Emergency Management of China.
The text structure of a Chinese accident case is listed in Fig. \ref{tree:structure}.
Some projects are small, and none of the stakeholders are corporations.
But the loss is severe.
So chapters about the project profile and accident prevention and improvements are sometimes omitted.
However, the accident causes are the core of the document.

Moreover, there are causes-and-effects relationships between two kinds of causes in the cases.
Direct causes have two main factors, which are unsafe behaviors of people and hazard status of matters.
Furthermore, the matter includes equipment, material, and surroundings.
The indirect cause can lead to immediate causes and thus increase the risk of projects, which is similar to distal causal in \citep{suraji2001development}.
Details of the indirect cause will be discussed in section \ref{sec:factor}.

\begin{figure}[ht]
\centering
  \includegraphics[width =0.68\textwidth]{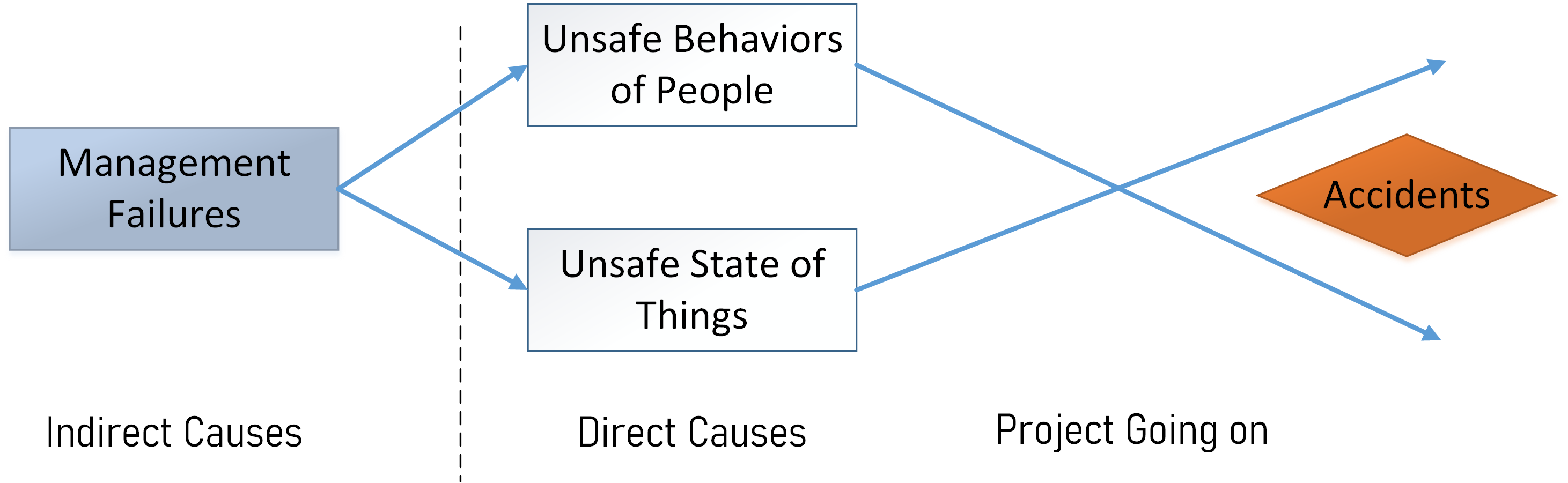}
\caption{Cross track model: causes-and-effects relationships between direct causes and indirect causes.}\label{fig:Cross_track}
\end{figure}

As discussed above, most of the Chinese cases are from accident investigation reports.
And the legal base of investigation reports is Regulation for the Investigation of Casualty Accidents of China (RICAC) \citep{China1986}.
Professor Sui is one of the counselors of RICAC, who proposed an accident model called the cross-track model \citep{SuiCP1982en}.
This model illustrates the relation between direct and indirect causes.
In Fig. \ref{fig:Cross_track}, the unsafe behaviors and hazard statuses of matters are understood as a consequence of management failures.
Moreover, the accident is not an inevitable outcome.
But as the project goes on, loss expectation will increase until an accident happens.

\section{Extract causal factors for each accident}\label{sec:extract}

\begin{figure}[ht]
\centering
  \includegraphics[width =0.88\textwidth]{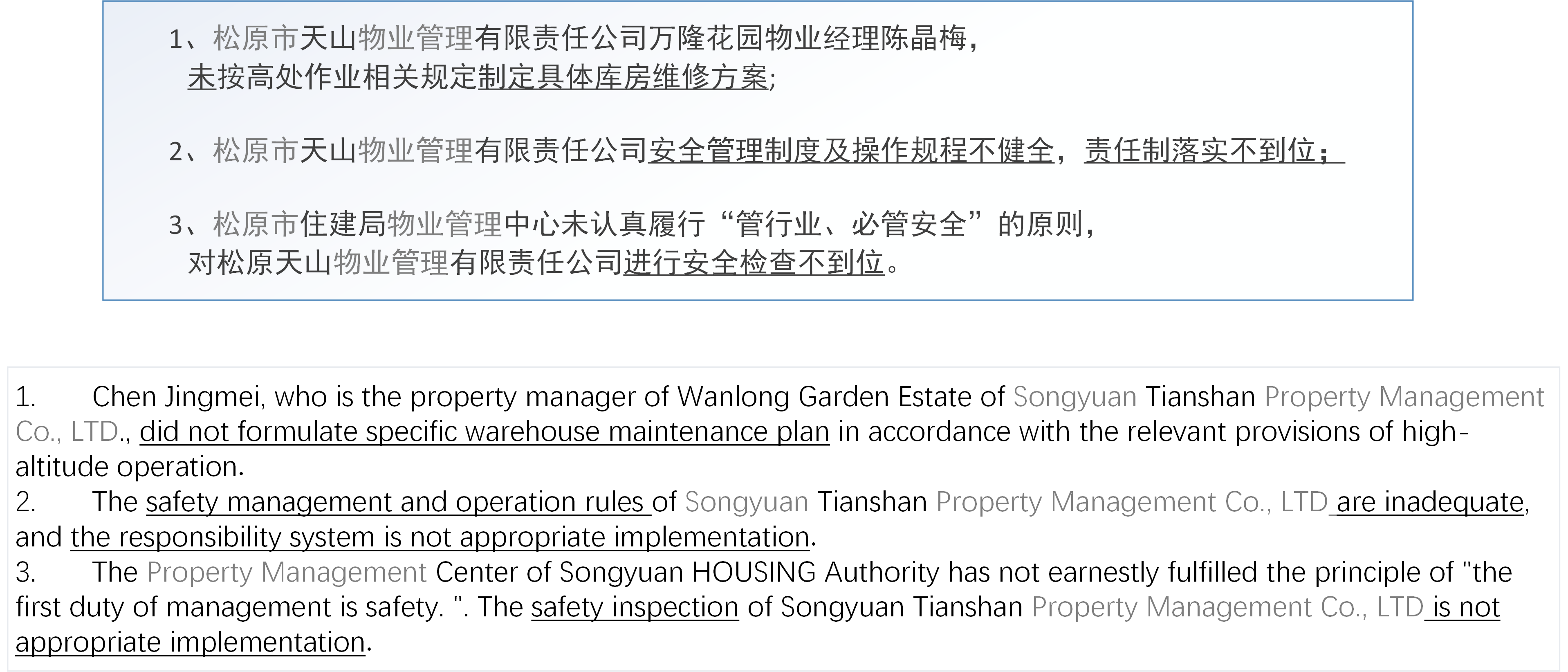}
\caption{An example for keyphrases, which are indirect causes of an accident. Texts with underline are the keyphrases for the cause. And the grey one is high-frequency word.}\label{fig:Case_text}
\end{figure}

Although causal factors are rich in the two specific sub-sections of case texts, not every word is about the cause.
In Fig. \ref{fig:Case_text}, there is an example from one real case text.
The parts with underlines describe the causes of the accident.
Rather than one single word, a sequence of words makes up this description, which is called the phrase \citep{merrouni2019automatic}.
Moreover, an observation in Fig. \ref{fig:Case_text} is that only some phrases are valuable to analyze causal factors.
In this paper, these phrases are called \texttt{keyphrases}.
Finally, more than one keyphrases can express the full meaning of accident causes.
This kind of keyphrases set is called the \texttt{fact}.

Each case text contains more than one fact about the accident.
Based on a series of NLP techniques, in this section, a framework will be proposed to extract these facts.
Due to the complexity and ambiguity of natural language, there are many ways of expressing the same semantic \citep{piskorski2013information}.
So it is almost impossible to find every fact from the free text.
Our study, due to the above, is based on one assumption that people and organizations repeat the same mistakes always.
As a result, if our framework can extract frequent causes automatically, the manual workload for the rests will be very reduced.

\subsection{Framework for extraction}
Automatic Keyphrase Extraction (AKE) is a task of natural language processing (NLP), which may be divided into two kinds \citep{merrouni2016automatic}: supervised and unsupervised.
Although promising results were delivered from current supervised AKE approaches, both the data labeling and manual sorting facts are time-consuming.
Without training data, unsupervised AKE is a recent trend aimed at discovering the underlying structure of a document \citep{alrehamy2017semcluster}.
The graph-based model is a typical method of unsupervised AKE \citep{washio2003state,sonawane2014graph}, in which the whole text is switch to the network, words as nodes.
Based on different standards, each node gets a weight to evaluate its importance.
Then rank nodes by their weight, and select nodes of top rank as keyphrases at last.
However, based on this graph-based model, it can not guarantee a phrase representing the text theme is a top-ranking term if it does not frequently occur in the text.
In the text of Fig. \ref{fig:Case_text}, the occurrences of some phrases are much higher than anyone of keyphrases.
For example, \emph{Songyuan} appears 3 times, \emph{Property Management} appears 4 time and \emph{Property Management Co., LTD.} is 3.

\begin{figure}[ht]
\centering
  \includegraphics[width =0.88\textwidth]{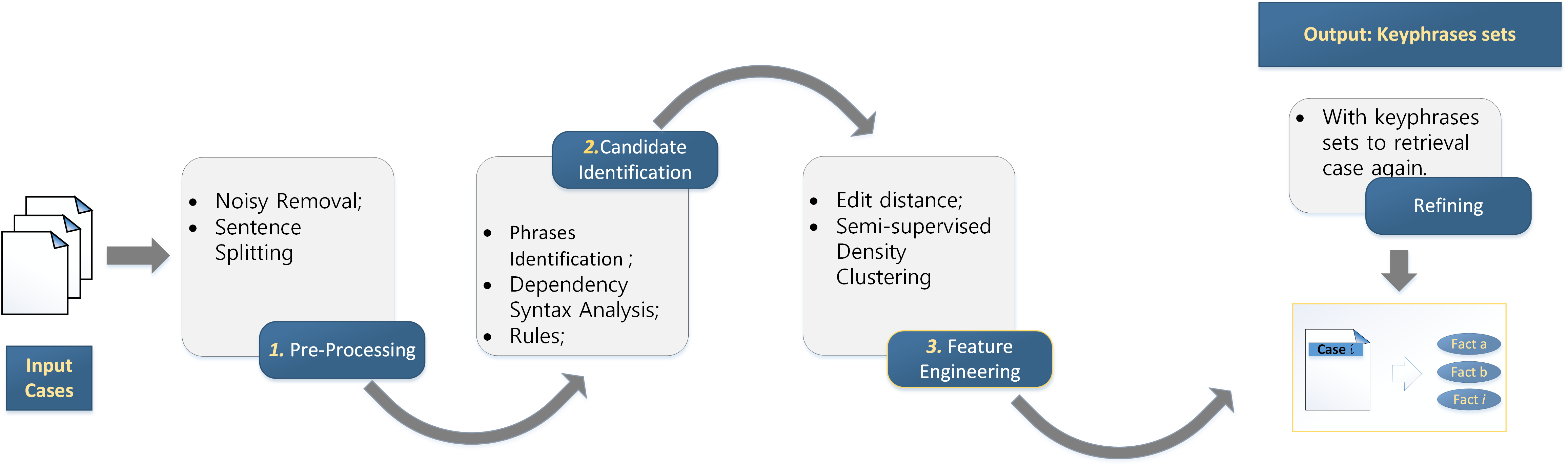}
\caption{Keyphrase extraction process stages.}\label{fig:flow}
\end{figure}

Although the weighted network's topology can not be used as an extraction basis in our data, the following two features are still valuable.
\begin{enumerate}
  \item
  The core of causal factors is usually verb phrases in Chinese.

  \item
  If people repeat the same mistake, which is the assumption discussed in section \ref{sec:extract}, one causal factor in one case will appear in others.
\end{enumerate}

Based on the above features about the case text, the whole workflow is depicted in Fig. \ref{fig:flow}.
In this process, the core parts are the candidate identification (step 2) and feature engineering (step 3).
In stage 2, candidate phrases sets will be identified through dependency syntax analysis (DSA) and heuristic rules.
The core meaning of every sentence will be extracted in this step.
If multiple candidate phrase sets have a similar semantic, in the next step, keyphrases sets (facts) will easily be brought together with the semantic clustering.

\subsection{Case text pre-processing}
In the preprocessing stage, text data will be formatted into a machine-readable format to decrease their complexity.
In Chinese, a part of a sentence that can provide additional information for the sentence is called the sense group \citep{ZhouC.2002}.
And sense groups of a sentence are divided into commas, semicolons, and full stops.
We believe that a sense group can retain the whole meaning of a fact.
To this end, after noisy symbols are removed, sentences will be segmented by the three kinds of punctuation.
In our studies, these segments are called candidate clauses.

\subsection{Identification of candidate phrases sets}
In this stage, candidate clauses will be transformed into candidate phrase sets.

For detecting all candidate phrases sets, three main methods were used by previous studies: N-Gram based \citep{huang2006keyphrase, liu2009clustering}, Part-Of-Speech (POS) sequence based \citep{barker2000using} and both \citep{grineva2009extracting}.
All methods above fall into the lexical analysis.
According to the characters of our data, a novel method based on syntactic analysis will be proposed in this paper.
Dependency parsing is quite a vital grammar analysis tool \citep{calvo2011dependency}.
In the dependency grammar, rather than the constituent and structure of solo phases, binary grammatical relations between words are directly described.

\begin{figure}[ht]
\centering
  \includegraphics[width =0.88\textwidth]{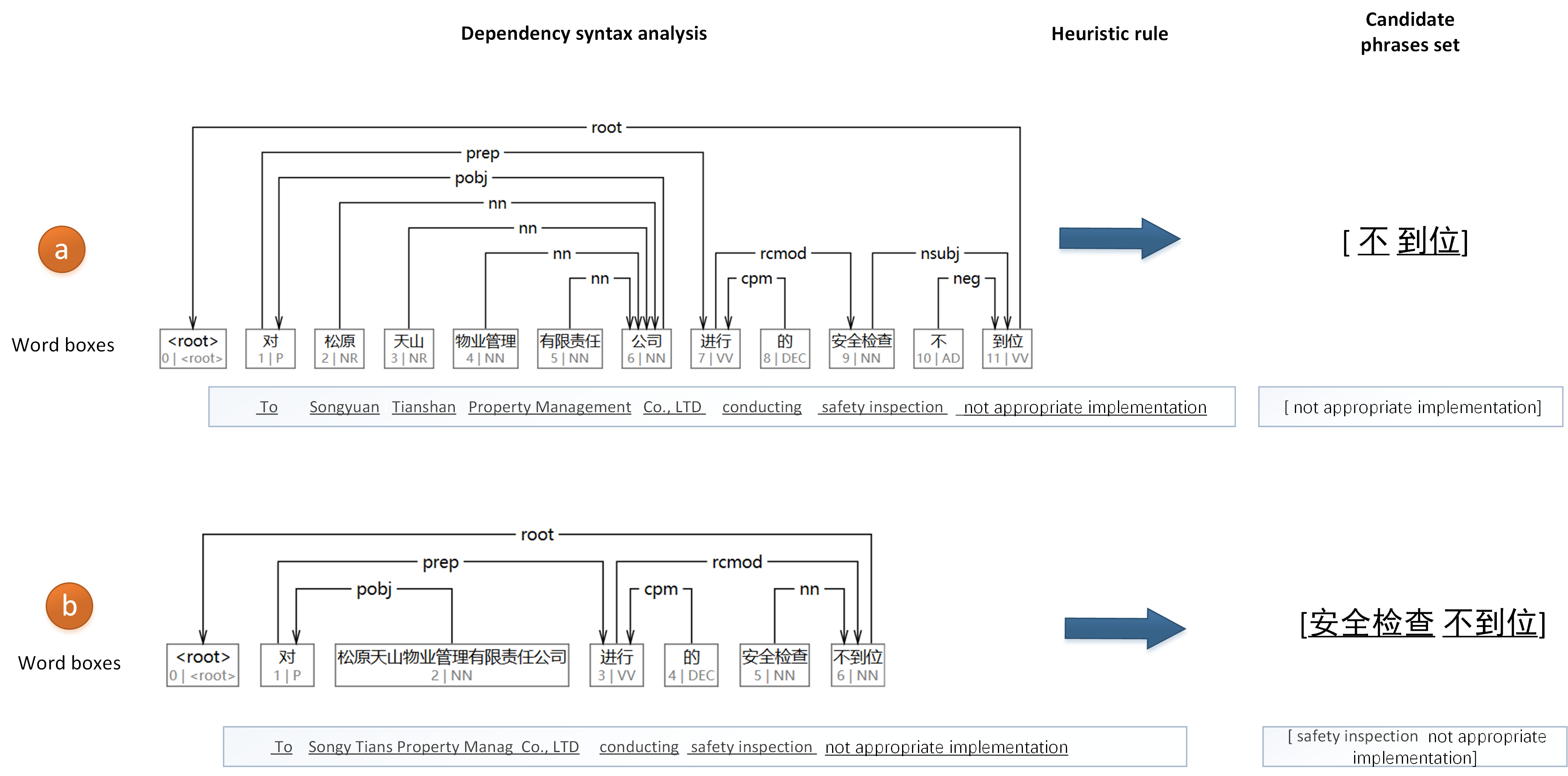}
\caption{Two examples of candidate identification. Candidate clauses in the two examples are the same. Because of different segmentations, results are very different. }\label{fig:dependency_ex}
\end{figure}

In Fig. \ref{fig:dependency_ex}, there are two examples of the sentence dependency syntax analysis (DSA).
A Chinese sentence is cut into words (or phrases).
Each one, in Fig. \ref{fig:dependency_ex}, is in a top part of a word box.
And the bottom part of this box is the sequence number and part of the speech of it.
On the top of the word boxes, the directed edge is from the headword to its dependent.
And the labels are all from a fixed library of syntax relations \citep{nivre2016universal}.
There must be a root node in the dependency structure, which is the head of others.
Note that if the sequence number of the headword is less than the dependent's, the arc is called \texttt{reverse} syntax relation.

With heuristic rules, the DSA results of candidate clauses will be extracted to get candidate phrase sets.
Generally, the root of a sentence is a verb phrase, which is the core of causal factors.
As a result, the start point of extraction of rules is the root phrase.
Further, the other two rules will help to find the rest candidate phrases if they exist.
\begin{enumerate}
  \item Taking the root as the start, its nearest dependent will be extracted.
  \item The headword and dependent of some certain reverse syntax relation will be extracted, which is the nearest to the root.
  These reverse syntax relations include direct object (\emph{dobj}), object of preposition (\emph{pobj}), adjectival complement (\emph{acomp}).
\end{enumerate}
In the sub-figure b of Fig. \ref{fig:dependency_ex}, the last phrase is the root of the whole sentence, so there is no reverse syntax relation in it.
And following rule 1, namely the nearest dependent of the root, safety inspection can be extracted.
By rules, the candidate phrases set of this clause is `safety inspection not appropriate implementation'.

Moreover, an important observation in Fig. \ref{fig:dependency_ex} is that although the candidate clauses and the rules are the same, the results are different.
The reason is different ways of sentence segmentation.
With classical methods, in sub-figure a of Fig. \ref{fig:dependency_ex}, the clause is divided into words.
And the complete fact can not be found by rules.
Rather than words, in the sub-figure b, the same sentence is cut into phrases.
The phrase, in Chinese, is a group of words or a single word, which is a single unit in the grammar of a sentence.
In the example above, a group of words is combined to \texttt{Songy Tians Property Manag Co., LTD} which is a noun phrase.
And the phrase in the last box is an adverbial phrase.

A few kinds of noun phrases, such as organizations and locations, can be found by one NLP technology called named-entity recognition (NER) \citep{nadeau2007survey}.
Other kinds of phrases, including some noun phrases, need a novel method.
Phrases extraction, essentially, is the assignment to identify combinations of words that show some idiosyncrasy in some certain corpus \citep{bouma2009normalized}.
In this paper, this idiosyncrasy will be evaluated by a mixed index \citep{Hankcs2019}.
The equation is as follows:

\begin{equation}\label{equ:score}
 Score(b) = PMI(w_i, w_j) + min( H_LC(b), H_RC(b) )
\end{equation}
In the phase extraction, two sequential words in the text are called the bigram.
Let $w_i, w_j$ be a bigram in the corpus, which is denoted by $b$.
The score of bigram $b$, in Equ. \ref{equ:score}, is composed of two parts, which will be used to evaluate whether $b$ can be a phrase.
Specifically, $PMI(w_i, w_j)$ is the inner connection index and $min( H_LC(b), H_RC(b) )$ is the outer independence index.

Pointwise mutual information (PMI) is one of the standard connection measures in the phrase extraction, which was introduced into NLP by Church and Hanks \citep{church1990word}.
\begin{equation}\label{equ:pmi}
 PMI(w_i, w_j) = \log(\frac{P(w_i, w_j)}{P(w_i)\times P(w_j)})
\end{equation}
$P(w_i, w_j)$ is the probability of the bigram $w_i, w_j$, which can be gotten by the maximum likelihood estimation.
$P(w_i, w_j) = C(w_i, w_j)/N$, where $C(w_i, w_j)$ is the number of occurrences of the bigram and $N$ is the number of words in the corpus.
By the same way, $P(w_i)$ and $P(w_j)$ can be estimated also.

PMI as an inner connection index can not be used to evaluate whether the bigram is a complete phase.
$PMI(Songyuan, Tianshan)$ \footnote{Songyuan is the name of a city. Tianshan is a mountain}, for example, may have a high PMI value.
But `Songyuan Tianshan Property Manag Co., LTD' is a whole noun phrase.
In other words, by the outer index, a bigram can be independent of contextual words.

If contextual words of a bigram are always in change, we believe that it may well be a complete semantic unit \citep{lee2019combining} (phrase).
Information entropy can be used to calculate the chaos and unpredictability of a random variable.
Let $LC(b) = \{w_1, ... , w_n\}$ be left context words set of the bigram.
Thus the left entropy of bigram can be defined as:
\begin{equation}\label{equ:HB}
 H_LC(b) = \sum_{w_i\in LC(b)}P(w_i)log_2 P(w_i)
\end{equation}
By MLE, $P(w_i) = C(w_i)/N$, where $C(w_i)$ is the number of occurrences of word $w_i$ appearing to the left of $b$, and $N$ is total number of occurrences that all adjacent words appear to the left of $b$.
In the same way, the right entropy of $b$ can also be got.

Finally, based on these scores, the bigrams set will be ranked.
And top-ranked ones may be returned as phrases.
Note that the phrase extraction can be operated repeatedly until as many whole semantic units as a possible return.

\subsection{Feature engineering}
In this step, accident facts will be identified.
In AKE, characters that can distinguish keyphrases from others in the candidate set are called features.
TF-IDF (Term frequency - Inverse document frequency) is the most popular feature \citep{nguyen2007keyphrase, liu2009unsupervised}.
TF-IDF can select candidate phrase sets that are frequent in a given document but infrequent in the whole corpus.
As shown in Fig. \ref{fig:Case_text}, facts can not be identified because of less frequency.
Assuming that people always repeat the same mistakes, a novel feature will help to pick keyphrases in our studies.

Repeating the same mistakes means the facts with similar semantics appear in many different candidate phrase sets.
As a result, the cluster based on the semantic similarity can characterize keyphrases sets from others.
By counting the minimum number of operations required to switch one string to the other, edit distance is a method to evaluate the semantic similarity \citep{baroni2002unsupervised} between two candidate phrases sets.
In our work, types of operation contain the insertion, removal, or substitution of a character in the string.
This kind of distance is called Levenshtein distance \citep{Navarro2001} which is defined as the following.
\begin{equation}\label{equ:distance}
 sem(a,b) = lev(a,b) = \left\{\begin{array}{lc}\vert a\vert&if\vert b\vert=0\\\vert b\vert&if~\vert a\vert=0\\lev(tail(a),tail(b))&if~\;a\lbrack0\rbrack=b\lbrack0\rbrack\\1+min\left\{\begin{array}{l}lev(tail(a),b)\\
 lev(a,tail(b))\\lev(tail(a),tail(b))\end{array}\right.&otherwise\end{array}\right.\\
\end{equation}

$lev(a,b)$ is the Levenshtein distance of the two strings $a, b$ and $|a|, |b|$ is the length of them.
The tail of string a ($tail(a)$) is the string of all but the first character of $a$, and $a\lbrack n\rbrack$ is the \emph{n}th character of the string \emph{a}, starting with character 0.
For the two strings $a, b$ ($|a|>0, |b|>0$), if they're exactly the same, $lev(a,b) = 1$.
Further, the larger the difference between $a, b$, the higher the Levenshtein distance.
As a result, $lev(a,b)$ can be used to evaluate semantic similarity.
Let $T_D$ be the candidate phrases.
Levenshtein distance is used to get the pairwise similarities between each pair of phases in $T_D$.
And the result is a similar matrix of size $|T_D| \times |T_D|$, which is denoted by $SC$.

Then, $SC$ will be clustered.
There are many kinds of algorithms to cluster $SC$ efficiently, but not all can analyze the distance matrix.
DBSCAN \citep{ester1996density} is a robust algorithm that does not need to specify the number of clusters.
DBSCAN requires two parameters.
One is the radius of a neighborhood with respect to some point denoted by $\varepsilon$.
The other is the minimum number of points ($minPts$) required to form a dense region.
A point is a core point if at least $minPts$ points (including the core point) are within distance $\varepsilon$ of it.
With the core point, DBSCAN will cluster all points (core or non-core) that are reachable from it.

\begin{figure}[ht]
\centering
  \includegraphics[width =0.68\textwidth]{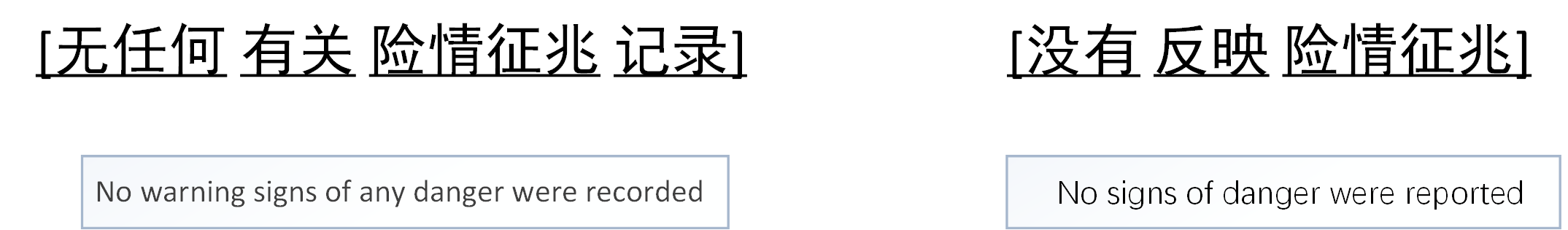}
\caption{The two sentences have similar semantics, which can be classified into one causal factor. However, the Levenshtein distance between them is not small, which is 1.9. }\label{fig:similarity_ex}
\end{figure}

Every parameter will influence the result of an algorithm, which is the key for every mining task.
To DBSCAN, $\varepsilon$ and $minPts$ as parameters are needed to specified by the user.
\begin{itemize}
  \item $minPts$ is then the desired minimum cluster size.
  Because people always repeat the same mistakes, $minPts$ can be set a little higher.
  Generally, higher values are better for data sets with noise sets and will yield more significant clusters.
  Here noise sets mean the content of the phrase set is nothing about the cause of the accident.
  In the clustering process of our study, $minPts$ is always 5.

  \item It is hard to estimate $\varepsilon$ because there are many ways to express the same semantic in the free text.
  But it is much easier to get a minimum value of $\varepsilon$ than its maximum value.
  If two candidate phrases sets are the same, which is very common in $SC$, the Levenshtein distance between them is 1.
  So the lower bound of $\varepsilon$ is 1.
  If $\varepsilon$ is chosen much too small, a large part of the data will not be clustered.
  The example is in Fig. \ref{fig:similarity_ex}.
  Two candidate phrase sets are all about warning signs being ignored when the Levenshtein distance between them is not small, which is 1.9.
  Namely, if $\varepsilon < 1.9$, it is quite possible that they are considered as noise set by DBSCAN.
  And for a too high value of $\varepsilon$, clusters will merge, and most nodes will be in the same cluster.
\end{itemize}

\begin{figure}[ht]
\centering
  \includegraphics[width =0.88\textwidth]{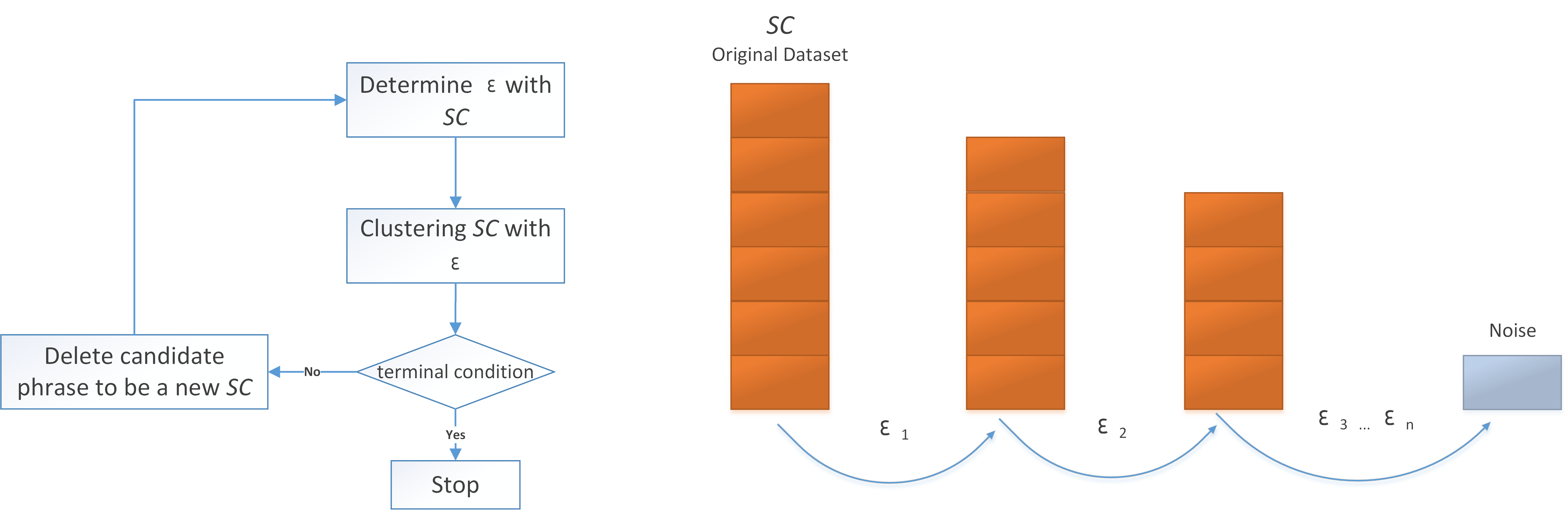}
\caption{The algorithm flow of multi-density. $\varepsilon_1 < \varepsilon_2 < ... <\varepsilon_n$. $SC$ will be clustered by $\varepsilon_1$. The the phrases in any clusters will be removed from the $SC$.  And the rests will be be clustered by $\varepsilon_2$. Repeat the two steps above until the condition is met.}\label{fig:multi_density}
\end{figure}

In our work, a succinct multi-density clustering will be implemented in our candidate phrases sets.
The algorithm is listed as the following:
\begin{enumerate}
  \item
  To candidate phrases set $SC$, $\varepsilon$ is determined by comparison.

  \item
  With $\varepsilon$, some clusters will be mined from $SC$.

  \item
  If any two phrases in one cluster satisfy $lev(a, b) = Max(|a|,|b|)$, the algorithm will stop.
  All clusters mined by the algorithm are the result.

  \item
  If not, delete candidate phrases set belonging to any clusters from $SC$ to form a new $SC$.
  And repeat step 1.

\end{enumerate}
The whole process is depicted in Fig. \ref{fig:multi_density}.

\begin{figure}[ht]
\centering
  \includegraphics[width =0.98\textwidth]{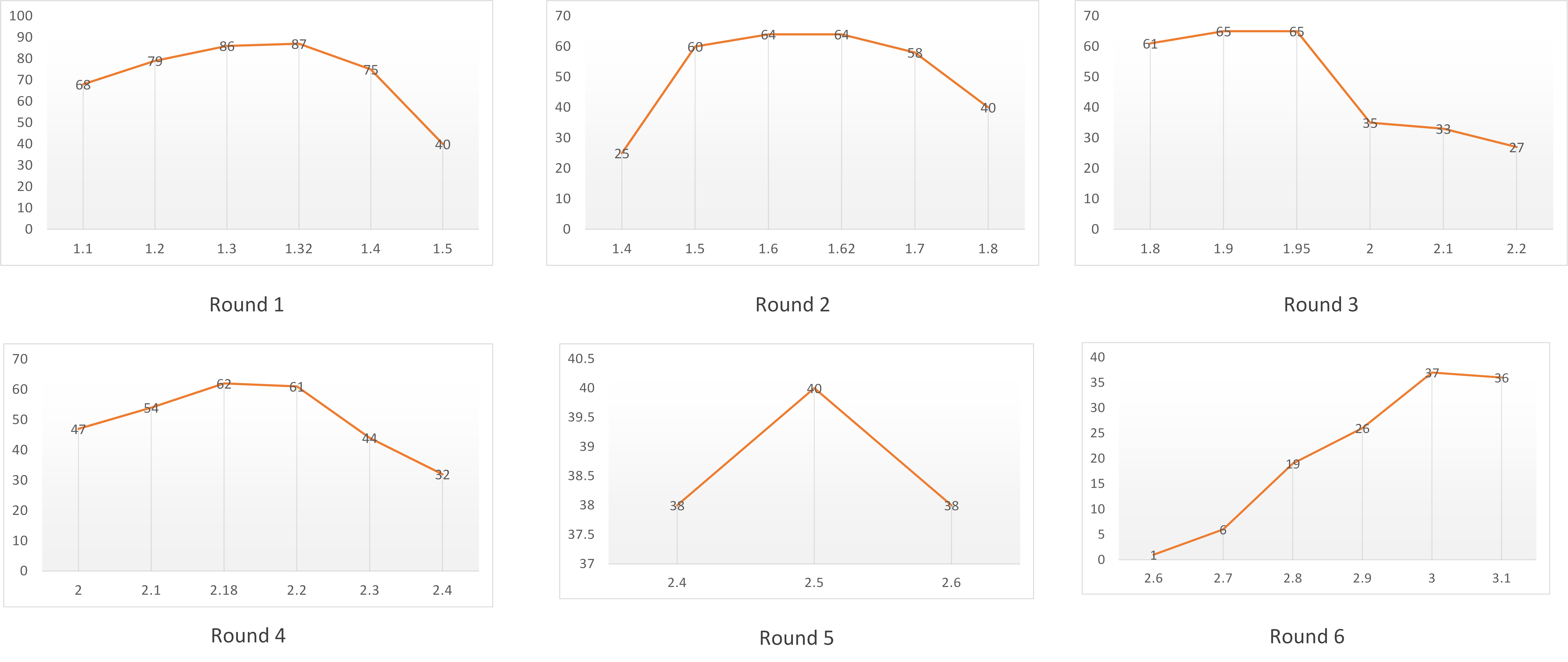}
\caption{The value of $\varepsilon$ used in the multi-density clustering.}\label{fig:episilon}
\end{figure}

The subgraph named Round1 in Fig \ref{fig:episilon} depicts the relationship between the $\varepsilon$ and the number of clusters.
The whole $SC$ is clustered by different $\varepsilon$ whose value is from 1.1 to 1.5.
The peak number of clusters appears in $\varepsilon = 1.32$, which is chosen as the value of the radius of a neighborhood in round 1.
The same pattern about the number of clusters appears in the rest of the data until the stopping rule is satisfied.
Note that the terminal rule is $lev(a, b) = Max(|a|,|b|)$, which means there is not one same character in the string $a$ and $b$.
In our data set, round 6 is the last clustering and $\varepsilon_6 = 3$
The radius from $\varepsilon_1$ to $\varepsilon_5$ are depicted in Fig. \ref{fig:episilon}.

\subsection{Summary for extracting causal factors}
267 accident case texts are input into our extracting framework, in which accidents happened from 1998 to 2018.
And 5598 candidate clauses format from these text data.
Of course, 5598 candidate phrase sets are ready for clustering analysis by DSA and heuristic rules extracting.
After six rounds of multi-density DBSCAN, 355 clusters are the final result, and 664 phrases sets are not contained by any clusters.
In 664 sets, only 3 are not noise sets.

Note that only 40 clusters (in 355) are noise set also.
After removing duplications, 1669 phrase sets about accident causations are the keyphrase sets.
Then each case text will retrieval these key sets to get the recall.
More specifically, if a clause in the text includes a whole essential phrase set, the causal factor is identified.
The recall of our framework is 87\%.

\section{New causal factors discovery}\label{sec:discovery}
As discussed above, the scale of risk factors in the construction are much larger than the causal factors.
An excellent review \citep{mohammadi2018factors} investigated 55 previous papers, and  95 sub-factors are summarized into 13 main factors.
In contrast to risk factors, ConAC, which is a causal model, only considers Four main factors and 19 sub-factors \citep{winge2019causal}.
As a result, for revealing new causal factors, we try to classify 1669 facts into 95 sub-factors until someone can not be laid down.
If some of these neglected facts have common characteristics, we can say one novel causal factor is discovered.

\subsection{Role mismatch}\label{sec:factor}
The first one, which caught our attention, is a fact which is `fake many times to defraud franchise'.
Not only is this fact not classified into any 13 primary factors, but it makes me wonder what has happened in that accident.
Then we went back to read the case text and found that it was a complicated accident \footnote{\url{http://www.safehoo.com/item/157796.aspx}~~~~ Last open in 2021.01.15}.
In brief, to save money, a big project is masqueraded as a small one by lying to the government first.
Then the client finish jobs of the contractor, supervision, and engineering designer.
Because of the improper plan, insufficient strength of columns led to concrete formwork collapsing.
Seven people died, and over ten were injured in this accident.
It is impossible for respondents in the questionnaire or interview to admit such a severe crime.

Stakeholders are the organizations who are actively involved with the project's work or have something to either gain or loss due to the project \citep{newcombe2003client}.
Much more than other industries, there are five kinds of directly involved organizations in China, including the government, client, project supervision, contractor, and others (Land survey, design, equipment leases, etc.).
One stakeholder unfulfilling his responsibility to result in an accident has drawn attention from previous studies  \citep{chen2013multilevel, pinto2011occupational}.
But few people note that one stakeholder did something beyond their scope of duties and cause accidents.
In this paper, this is called \textbf{role mismatch}.
One example is the client in the last paragraph.

From 267 case texts, six kinds of role mismatch are summarized, which is listed in Tab. \ref{tab:mismatch}.
Except for supervisors, the other five kinds of stakeholders are included.
The second column of Tab. \ref{tab:mismatch} is the occurrence number of this sub-factor in total 267 cases.
If two factors appear in the same accident frequently, there may be a strong correlation between them.
Before discussing the relations between role mismatch and other causal factors, a classification of causal factors will be proposed first.

\begin{table}[htbp]
\centering
 \caption{Relations between role mismatch and other causal factors }\label{tab:mismatch}
 \vspace{0.5cm}
 \scriptsize
 \begin{tabular}{p{4cm}p{2cm}<{\centering}p{2cm}<{\centering}p{2cm}<{\centering}p{2cm}<{\centering}p{2cm}<{\centering}}
   \hline
    Role mismatch & Occurrence number   & \multicolumn{4}{c}{Other factors in the same accident} \\
   \hline
   \hline
       Client: making construction plan & \textbf{1} & 	\texttt{6-2}(\textbf{1})	&	\texttt{2-7}(\textbf{1})	&	\texttt{2-12}(\textbf{1})	&	\texttt{2-3}(\textbf{1})\\
    \hline
    Government: appoint sub-contractor& \textbf{1} &	\texttt{2-12}(\textbf{1})	&	\texttt{2-7}(\textbf{1})	&	\texttt{3-1}(\textbf{1})	&	\texttt{2-3}(\textbf{1})\\
    \hline
  & &  \texttt{2-7}(\textbf{33}) &	\texttt{2-12}(\textbf{12}) &	\texttt{3-4}(\textbf{5})&  \texttt{4-1}(\textbf{16})\\
  & &  \texttt{2-13}(\textbf{26}) &	\texttt{1-3}(\textbf{11}) &	\texttt{5-2}(\textbf{3})&	\texttt{2-11}(\textbf{6}) \\
   Contractor: construction &\textbf{41} &  \texttt{2-3}(\textbf{23}) &	\texttt{3-1}(\textbf{11}) &	\texttt{4-3}(\textbf{2})&	\texttt{3-5}(\textbf{1})\\
  without competency& &  \texttt{1-1}(\textbf{23}) &	\texttt{2-4}(\textbf{11}) &	\texttt{6-1}(\textbf{2})&  \texttt{2-5}(\textbf{15})\\
  &  &  \texttt{5-1}(\textbf{22}) &	\texttt{2-8}(\textbf{9}) & \texttt{2-9}(\textbf{2})&	\texttt{6-2}(\textbf{6})\\
  & &  \texttt{1-2}(\textbf{18}) &	\texttt{2-1}(\textbf{9}) &	\texttt{4-4}(\textbf{2})&  \texttt{2-10}(\textbf{13})\\
  & &  \texttt{3-2}(\textbf{16}) &	\texttt{2-6}(\textbf{8}) &	\texttt{4-2}(\textbf{1})&	\texttt{2-2}(\textbf{5})\\
  \hline
    & &  	\texttt{2-13}(\textbf{5})			 &	\texttt{1-1}(\textbf{3})			 &	\texttt{1-2}(\textbf{2})	&  	\texttt{3-2}(\textbf{3}) 		\\
    & &  	\texttt{2-7}(\textbf{5})			 &	\texttt{2-10}(\textbf{3})			 &	\texttt{3-1}(\textbf{2})	&	\texttt{4-3}(\textbf{1})		\\
   Contractor:  & \textbf{5}&  	\texttt{2-3}(\textbf{5})			 &	\texttt{2-5}(\textbf{2})			 &	\texttt{2-6}(\textbf{1})		&	\texttt{4-1}(\textbf{2})	\\
 illegal transfer&  &  	\texttt{5-1}(\textbf{4})			 &	\texttt{2-11}(\textbf{2})			 &	\texttt{1-3}(\textbf{1}) &	\texttt{2-12}(\textbf{2})			\\
& &  	\texttt{2-8}(\textbf{3})			 &	\texttt{5-2}(\textbf{2})			 &	\texttt{2-2}(\textbf{1})&	\texttt{2-4}(\textbf{1})			\\
    & &  	\texttt{2-1}(\textbf{3})						 			\\

  \hline
    &  &	\texttt{1-2}(\textbf{48})			&	\texttt{2-6}(\textbf{22})			&	\texttt{4-3}(\textbf{7})	&	\texttt{2-12}(\textbf{27})		\\
    &  &	\texttt{2-3}(\textbf{43})			&	\texttt{1-3}(\textbf{19})			&	\texttt{3-4}(\textbf{6})	&	\texttt{2-1}(\textbf{12})		\\
    &  &	\texttt{2-7}(\textbf{40})			&	\texttt{2-5}(\textbf{19})			&	\texttt{2-11}(\textbf{5})	&	\texttt{4-2}(\textbf{2})		\\
   Worker: labour & \textbf{57} &	\texttt{5-1}(\textbf{37})			&	\texttt{2-10}(\textbf{18})			&	\texttt{6-2}(\textbf{5})	&	\texttt{2-4}(\textbf{25})		\\
   without competency &   &	\texttt{4-1}(\textbf{35})			&	\texttt{2-13}(\textbf{17})			&	\texttt{6-1}(\textbf{5})			&	\texttt{3-1}(\textbf{12})\\
    &  &	\texttt{1-1}(\textbf{31})			&	\texttt{3-2}(\textbf{16})			&	\texttt{5-2}(\textbf{4})			&	\texttt{4-4}(\textbf{2})\\
    &  &	\texttt{2-8}(\textbf{30})			&	\texttt{2-2}(\textbf{14})			&	\texttt{2-9}(\textbf{3})			\\

    \hline
    &  &	\texttt{2-7}(\textbf{4})			&	\texttt{1-2}(\textbf{3})			&	\texttt{6-1}(\textbf{1})			&	\texttt{4-4}(\textbf{1})			\\
    Designer: & \textbf{4} &	\texttt{5-1}(\textbf{4})			&	\texttt{1-1}(\textbf{3})			&	\texttt{2-2}(\textbf{1})			&	\texttt{2-10}(\textbf{1})			\\
    without competency&  &	\texttt{6-2}(\textbf{4})			&	\texttt{3-2}(\textbf{3})			&	\texttt{2-12}(\textbf{1})			&	\texttt{3-4}(\textbf{1})			\\
    &  &	\texttt{2-13}(\textbf{4})			&	\texttt{2-4}(\textbf{2})			&	\texttt{2-1}(\textbf{1})			&	\texttt{3-5}(\textbf{1})			\\
    &  &	\texttt{2-3}(\textbf{4})			&	\texttt{3-1}(\textbf{2})			&	\texttt{4-1}(\textbf{1})			\\				\hline

 \end{tabular}
\end{table}

The case data character is each accident fact has a stakeholder who has to be held accountable.
As a result, six main factors correspond to six different kinds of stakeholders in the construction industry in our classification.
Each stakeholder's responsibilities in the construction safety are defined in two laws of China \citep{PeoplesRepublicofChina2000, NationalPeoplesCongress2019}, which are Construction Law and Regulations on construction engineering quality management, respectively.
So the sub-factors are all from the two laws.
Rather than open interpretation \citep{behm2013application}, the definition of these factors in the law is more strict.
The main factors and their sub-factors are listed in Tab. \ref{tab:causal_factors} of Appendix A.
Note that the number in the bracket is the code of this sub-factor.
And these codes correspond to the number in the last column in Tab. \ref{tab:mismatch}.

If two factors appear in the same accident frequently, there may be a strong correlation between them.
Moreover, the causal diagrams \citep{pearl2018book} of the construction accident can be deduced from these correlations.
With role mismatch, factors that appear in the same accident are listed in the last column in the Tab. \ref{tab:mismatch}.
And the number of co-occurrence is in the bracket behind the factor code.
Based on causal diagrams \citep{pearl2018book}, the mechanism of role mismatch will be discussed in our future work.
Here we only come up with some preliminary observations.
Except for government appointing sub-contractor, reducing costs and saving time may be the common purpose of the rest five sub-factors.

\subsection{More than one neglected factor}
\begin{table}[htbp]
\centering
 \caption{The other two neglected causal factors}\label{tab:neglected}
 \vspace{0.5cm}
 \scriptsize
 \begin{tabular}{p{2.5cm}p{5cm}p{8cm}}
   \hline
    Main factor  & Sub-factors & Accident case title\\
   \hline
   \hline
   & Supplier: Failure to fully perform the contract &  2003-9-20 Lift cage falling\\
   & No engineer contract   & 1996-3-14 The earth collapsed\\
   Engineer contract management & No labor contract & 2003-5-15 The car crane collided with the high voltage line\\
   & In Inappropriate contract management & 2002-3-15 Crane boom overturned\\
   &  & 2003-9-12 Pipe network trench collapse\\
   &  & 2002-11-6 Falling\\

   \hline
   & Delayed response & 2001-6-20 The outer cornice collapsed\\
   &  & 2003-7-24 The building collapsed\\
   & No contingency plan & 2002-5-12 Explosion \\
    Response &  & 2003-11-20 Construction collapse\\
   for the accident& Contingency plan has not been implemented & 2014-9-1 Poisoning in a sewage pumping station project\\
   & Inappropriate rescue &  2003-3-29 Poisoning in a sewerage project\\
   &   &  2012-12-23 Carbon monoxide poisoning\\
   \hline
 \end{tabular}
\end{table}

From case text data, the other two main factors are relatively little studied in construction accidents.
One is engineering contract management, and the other is the response to the accident.
We believe that the reason for neglecting is also data problems.
It is hard to collect enough samples because the people who have participated in rescue or contract management are very few.

With case texts, other scholars may be inspired by the two factors and their sub-factors.
In Tab. \ref{tab:neglected}, the sub-factors are listed in the second column.
Note that all of these sub-factors are summarized from the real accident cases, and the date and title of them are in the last column.
And our share data in Github has these case texts.

\section{Conclusion}
The accident data is valuable.
After the whole process of past accidents is revealed, the future losses can be reduced.
Very few people have ever had an accident, so the data about accidents are hard to get.
Typical accident cases should be studied carefully because the cost of life is behind most of these texts.
Beyond the limitations of the manual analysis, based on a series of NLP technologies, a framework to organize data about accident causes is proposed in this paper.
And some neglected causal factors are discovered.
Role mismatch will be further discussed in our future studies.

We believe that our framework can also analyze Chinese case texts in other industries.
And the research involving other languages can be inspired by this work.
Moreover, society and economic climate can also affect the occupational incident system \citep{winge2019causal, chen2013multilevel}.
As a result, other developing countries would benefit from our study also.

\begin{flushleft}
\textbf{Acknowledgements}
\end{flushleft}
This work is supported by the National Natural Science Foundation of China Nos. 71501022, 71901047, 71874020 and 71774021.

\section*{Appendix A}

\begin{table}[htbp]
\centering
 \caption{Causal factors categorized by stakeholders}\label{tab:causal_factors}
 \vspace{0.5cm}
 \scriptsize
 \begin{tabular}{p{4cm}p{11cm}}
   \hline
    Stakeholder  & Causal factors (\texttt{ID})\\
   \hline
   \hline
        & $\bullet$  Unsafe operation (\texttt{1-1})\\
       1. Worker and Work group & $\bullet$  Without competency (\texttt{1-2})\\
        & $\bullet$  Tacit knowledge: ability, experience, knowledge, safety awareness (\texttt{1-3})\\
   \hline
    & $\bullet$   Responsibilities of contractor is not fulfilled (\texttt{2-1})\\
    & $\bullet$  Construction plan (\texttt{2-2})\\
    & $\bullet$  Safety, quality supervision and control (\texttt{2-3})\\
    & $\bullet$  Rules and regulation  (\texttt{2-4})\\
    & $\bullet$  Safety culture and climate  (\texttt{2-5})\\
   2. Contractor & $\bullet$  Safeguard procedures, equipment and sign  (\texttt{2-6})\\
    & $\bullet$  Inappropriate construction operation  (\texttt{2-7})\\
    & $\bullet$  Training and education  (\texttt{2-8})\\
    & $\bullet$  Site condition  (\texttt{2-9})\\
    & $\bullet$  Command  (\texttt{2-10})\\
    & $\bullet$  Verification of competency (\texttt{2-11})\\
    & $\bullet$  Response to the accident (\texttt{2-12})\\
    & $\bullet$  Competency of itself  (\texttt{2-13})\\
   \hline
    & $\bullet$  Safety management  (\texttt{3-1})\\
    & $\bullet$  Illegal construction  (\texttt{3-2})\\
   3. Client & $\bullet$ Supervising contractors  (\texttt{3-3})\\
    & $\bullet$  Project acceptance  (\texttt{3-4})\\
    & $\bullet$  Archives management   (\texttt{3-5})\\
   \hline
    & $\bullet$  Supervising contractors  (\texttt{4-1})\\
   & $\bullet$  Communication with client (\texttt{4-2})\\
   4. Supervisor& $\bullet$ Competency of itself   (\texttt{4-3})\\
   & $\bullet$ Tacit knowledge: ability, experience, knowledge, safety awareness  (\texttt{4-4})\\
   \hline
   & $\bullet$  Guide and supervise  (\texttt{5-1})\\
   5. Government & $\bullet$  Inappropriate punishment (Punishment is too light or laws is not strictly enforced) (\texttt{5-2})\\
   & $\bullet$  Organization, mechanism, system (\texttt{5-3})\\
   \hline
   6. Others  & $\bullet$  Supplier: Material and equipment quality (\texttt{6-1})\\
     & $\bullet$ Designer: Survey and design  (\texttt{6-2})\\
   \hline
 \end{tabular}
\end{table}

\newpage

\section*{Reference}

\bibliographystyle{elsarticle-num}
{\bibliography{d:/==-studies-==/reference}}

\end{document}